\documentclass[runningheads]{llncs}
\usepackage{graphicx}
\usepackage{booktabs}
\usepackage{amsmath, amssymb}
\usepackage{multirow}

\DeclareMathOperator{\Var}{Var}
\DeclareMathOperator{\E}{E}

\begin{document}
\title{Neural Networks with Different Initialization Methods for Depression Detection}
\titlerunning{Neural Networks for Depression Detection}
\author{Tianle Yang}
\institute{School of Computing in the College of Engineering and Computer Science, Australian National University, Australia \\
\email{u6512077@anu.edu.au}}
\maketitle              
\begin{abstract}

As a common mental disorder, depression is a leading cause of various diseases worldwide. Early detection and treatment of depression can dramatically promote remission and prevent relapse. However, conventional ways of depression diagnosis require considerable human effort and cause economic burden, while still being prone to misdiagnosis. On the other hand, recent studies report that physical characteristics are major contributors to the diagnosis of depression, which inspires us to mine the internal relationship by neural networks instead of relying on clinical experiences. In this paper, neural networks are constructed to predict depression from physical characteristics. Two initialization methods are examined - Xaiver and Kaiming initialization. Experimental results show that a 3-layers neural network with Kaiming initialization achieves $83\%$ accuracy.

\end{abstract}
\section{Introduction} \label{sec:introduction}
Clinical depression is a psychotic emotional disorder, mainly caused by the individual’s difficulty in coping with stressful life events. Depression negatively affects the patients, causes feelings of extreme sadness, and leads to various mental and physical diseases\cite{cite-2}. Depression is recognized as one of the risk factors for suicide \cite{cite-1}. The World Health Organization (WHO) ranks it as the fourth leading cause of disability in the world and predicts that it will become the second leading cause of disability by 2030.

As depression becomes common in the general population \cite{cite-4} and a major burden for the healthcare system worldwide \cite{cite-6}, effective depression diagnosis and treatment techniques attract increasing attention. However, early diagnosis of depression is clinically challenging. The diagnosis of depression is mostly given by general practitioners. Unfortunately, the modest prevalence of depression in primary care indicates that misidentifications outnumber missed cases\cite{cite-6}. 

Recently, depression diagnosis based on critical behavioral signals and physiological indicators is gaining growing popularity\cite{cite-7}. The study of objective biological, physiological and behavioral markers not only improves the accuracy of psychological diagnosis and treatment of many mental diseases but also eases the social and economic burdens associated with these diseases\cite{cite-8}. According to recent studies, there is an internal relationship between physical characteristics and the risk of depression. However, the relationship is so complicated that beyond current clinical experiences. Therefore, a model that learns the relationship from physical data to predict the diagnosis results is of great importance.

Machine learning is a powerful data analysis tool prevalent in both academia and industry. Studies in recent years have found it feasible and effective in the illness diagnosis. For example, decision tree-based classifiers are employed in the discovery of type \uppercase\expandafter{\romannumeral2} diabetes\cite{cite-12}. Ahmad\cite{cite-11} applies decision tree classifiers to diagnose breast cancer. Recent literature\cite{cite-9} uses neural networks in the field of predicting depression\cite{cite-9} and reports dramatically better results than human efforts.

In this paper, we emphasize neural networks for depression diagnosis based on observed behavioral signals and critical physiological indicators. Similar data preprocessing methods as in\cite{cite-9} are adopted. Without prior knowledge concerning depression, physical data from 12 individuals (6 men and 6 women) are collected while they are watching videos, including galvanic skin response (GSR), skin temperature (ST), and pupillary dilation (PD). Galvanic skin response exhibits unique patterns that indicate the response of sweat glands to different emotional stimuli. Skin temperature reveals the intensity of acute stress that the individual feels. Pupillary dilation provides signs of changes in mental activity intensity, and pupil size varies with emotional stimuli. In the experiment, we obtained 23 GSR features, 39 PD features, and 23 ST features in total. These features are processed and used to train the neural networks with different layers. We also investigate two typical initialization methods - Xaiver and Kaiming initialization. Experimental results show that a 3-layers neural network with Kaiming initialization achieves $83\%$ accuracy, which is the highest among all configurations experimented.

\section{Method} \label{sec:method}
In this paper, we adopt neural network classifiers and emphasize the effect of different layer numbers and initialization methods. Two initialization methods are under experiment: Xaiver\cite{cite-13} and Kaiming\cite{cite-10} initialization. Then, we investigate different configurations by varying the initialization methods and the number of layers. The optimal hyper-parameters for each model is chosen through experiments.

\subsection{Neural Network}
Artificial neural networks are hotspots in many fields. A neural network is composed of several layers, each of which contains multiple neurons. A neuron typically consists of three components: connection weight, adder, and activation function. Once many values from previous layers arrive at the input, the values are first multiplied by the weights on each connection. Then the adder sums up all the weighted values and forms the actual signal to the neuron. Finally, the activation function maps the input signal to a certain value within a permitted range. The dataset is preprocessed and sent to the first layer of the network. These values flow across layers and are processed by each layer. The final outputs of the last layer encode information for classifications.

Three kinds of neural network architectures are evaluated in the paper: single-layer, 2-layer, and 3-layer neural networks. For a 2-layer neural network, the number of hidden layers we use is 50, the same as in \cite{cite-9}. For a 3-layer neural network, the number of the first hidden layers is 50, and the number of the second hidden layer is 20. These parameters are explored and selected in the experiment. Neural network structures with deeper layers are not considered due to the limited samples.

In this paper, we mainly focus on how different initialization methods influence the prediction results. For all models, the learning strategy adopted is statistic gradient descent(SGD) with momentum. We explore and choose the optimal hyperparameters for multi-layer perception(MLP) networks. Table \ref{tab:parameter} lists the hyperparameters selected for different network models with Xaiver and Kaiming initialization, including batch size(bs), learning rate(lr), and momentum (m). These parameters are chosen to maximize the accuracy of a certain network structure with specific initialization methods so that we can compare the optimality between different structures and initialization methods.  

\begin{table}[h]
    \centering
    \caption{\textsc{Hyperparameters For Xaiver and Kaiming Initialization.}} \label{tab:parameter}
    \begin{tabular}{ccccccc}
        \toprule
                & \multicolumn{3}{c}{\textbf{Xaiver}} & \multicolumn{3}{c}{\textbf{Kaiming}} \\ 
                \cmidrule(r){2-4} \cmidrule(r){5-7}
                & bs & lr     & m   & bs & lr     & m \\
        \midrule
        1-layer & 24 & 0.0001 & 0.6 & 36 & 0.0001 & 0.6 \\
        2-layer & 24 & 0.006  & 0.7 & 36 & 0.003  & 0.7 \\
        3-layer & 36 & 0.006  & 0.7 & 36 & 0.0002 & 0.6 \\
        \bottomrule
    \end{tabular}
\end{table}

\subsection{Xaiver Initialization}

\noindent The parameters need to be initialized before neural network training. Weight initialization are typically randomized based on Gaussian distribution. 
However, with neural network depth increases, this approach suffers dramatically from gradient disappearance. The variance of activation values can be decreased layer by layer, 
causing the gradient vanishing layer by layer in the back propagation process. For training deeper neural networks, it is necessary to avoid the attenuation of the variance of the activation value.
To tackle this problem, Xaiver Glorot\cite{cite-13} proposes that the output value of each layer should keep Gaussian distribution in both forward and backward propagation, 
which is the core of the Xaiver initialization method.

A forward propagation involves the following calculations:
\begin{gather}
    \begin{gathered}
        Y_i = W_iX_i + B_i \\ 
        w \in \mathbb{R}^{u\times d}, x \in \mathbb{R}^d, b\in \mathbb{R}^u
    \end{gathered}
\end{gather}
\noindent where $Y_i, W_i, X_i, B_i$ correspond to the outputs, weights, inputs, and biases of the neurons in the $i^{th}$ layer.
In order to keep the forward signal strength unchanged, a necessary condition is to meet the requirements:
\begin{equation}
    \Var(Y_i) = \Var(X_j)
\end{equation}
Based on the following assumption of variable distribution:
\begin{gather}
    W,X,B \ \text{are independent of each other} \notag \\
    W_{ij} \ \text{i.i.d. and } \E[W_{ij}] = 0 \\ 
    B_{i} \ \text{i.i.d. and } \Var(W_{ij}) = 0 \\
    X_{j} \ \text{i.i.d. and } \E[X_{j}] = 0
\end{gather}
We can get:
\begin{gather}
    \begin{aligned}
    \Var(Y_i) &= \Var(W_iX+B_i) \\
              &= \Var\left( \sum_{j=1}^dW_{ij}X_j+B_i \right) \\
              &= d\times \Var(W_{ij}X_j) \\
              &= d\times(\E[W_{ij}^2]\E[X_j^2] - \E^2[W_{ij}]\E^2[X_j]) \\
              &= d\times \Var(W_{ij})\Var(X_j)
    \end{aligned}
\end{gather}

In order to guurantee $\Var(Y_i)=\Var(X_j)$, we need to satisfy $d\times \Var(W_{ij}) = 1$, which is equivalent to:
\begin{equation}
    \Var(W_{ij}) = \frac{1}{d}
\end{equation} 
Finally we get the following Xaiver initialization method:
\begin{itemize}
    \item For the normal distribution, $W_{ij} \sim \textrm{Normal}\left(0, \frac{1}{d} \right)$
    \item For the uniform distribution, $W_{ij} \sim \textrm{Uniform}\left( -\sqrt{\frac{3}{d}}, \sqrt{\frac{3}{d}} \right)$
\end{itemize}

\subsection{Kaiming Initialization}
Although Xaiver takes the variance of activation value into account, the activation function is still possible to change the distribution of the values flowed across layers.
Kaiming initialization\cite{cite-10} is proposed to solve this problem.

Consider a forward propagation:
\begin{gather}
    Z = f(X) \\
    Y = WX + B \\
    f \text{is ReLu function}, w \in \mathbb{R}^{u\times d}, x,z \in \mathbb{R}^d, y,b\in \mathbb{R}^u \notag
\end{gather}

Based on Xaiver initialization, a new hypothesis is introduced in Kaiming initialization: $X_j$ has a symmetric distribution around 0, which means:
\begin{equation}
    \Var(X_j) = \frac{1}{2}\Var(X_j)
\end{equation}
And $Var(Y_i) = Var(X_j)$ is still satisfied. Put it altogether, the Kaiming initialization method is as follows:
\begin{itemize}
    \item For the normal distribution, $W_{ij} \sim \textrm{Normal}\left(0, \frac{2}{d} \right)$
    \item For the uniform distribution, $W_{ij} \sim \textrm{Uniform}\left( -\sqrt{\frac{6}{d}}, \sqrt{\frac{6}{d}} \right)$
\end{itemize}

\section{Results and Discussion}\label{sec:results}
\subsection{Evaluation Metrics}
Based on the method discussed in Sec. \ref{sec:introduction}, a total of 192 pieces of data are collected from 16 participants and each of them own 12 records. Among the total dataset, 
$20\%$ of the data is set as the test set. For the remaining data, leave one out method is adopted to divide the training dataset and the validation set.

The metrics to measure the performance of different models are accuracy, precision, recall and F1 score. 
In the binary classification problem, it is assumed that the sample has two categories: positive and negative. Depending on the prediction result, all samples are classified into 4 classes:
\begin{itemize}
    \item True positive(TP): positive samples predicted to be positive. 
    \item True negative(TN): negative samples predicted to be negative. 
    \item False positive(FP): positive samples predicted to be negative.
    \item False negative(FN): negative samples predicted to be positive.
\end{itemize}
The definitions of accuracy, precision, recall and F1 score are as follows:
\begin{gather}
    \text{Accuracy} = \frac{TP+TN}{TP+TN+FP+FN} \\
    \text{Precision} = \frac{TP}{TP+FP} \\
    \text{Recall} = \frac{TP}{TP+FN} \\
    \text{F1 score} = \frac{2\times \text{Precision} \times \text{Recall}}{text{Precision} + \text{Recall}}
\end{gather}

\subsection{Results}
We use the dataset to train the network configurations with different number of layers and initialization methods as stated in Sec. \ref{sec:method}, using parameters listed in table \ref{tab:parameter}. The experimental results are listed in table \ref{tab:res-1} - \ref{tab:res-3}.

As depicted in table \ref{tab:res-1}, with Kaiming initialization, the precision and recall are both 0 for the moderate class, which signifies that the performance of the model is extremely poor. From the 3 tables listed, we can draw several conclusions.
\begin{enumerate}
    \item For both initialization methods, increasing the number of layers in the neural network helps to increase the overall performance. 
    \item For all network topologies, the initialization method dramatically affects the final accuracy results.
    \item While the performance of Xaiver remains relatively stable, performance of Kaiming initialization improves faster as the number of layers increases. It performs much worse than Xaiver in the 1-layer network, comparable in the 2-layer network, and better in the 3-layer network. Therefore, Kaiming initialization is more sensitive to network topologies.
    \item 3-layer neural network with Kaiming initialization achives the best accuracy of $83\%$ among all configurations.
\end{enumerate}

\begin{table}[h]
    \centering
    \caption{\textsc{Results for 1-Layer Depression Recognition Models.}} \label{tab:res-1}
    \resizebox{\textwidth}{!}{
    \begin{tabular}{ccccccc}
        \toprule
        \multirow{2}{*}{\textbf{Depression Level}} & \multicolumn{3}{c}{\textbf{1-layer+Xaiver}} & \multicolumn{3}{c}{\textbf{1-layer+Kaiming}} \\
                \cmidrule{2-4} \cmidrule{5-7}
                & \textbf{Precision} & \textbf{Recall} & \textbf{F1 score} & \textbf{Precision} & \textbf{Recall} & \textbf{F1 score} \\
        \midrule
                None & 0.38 & 0.36 & 0.37 & 0.89 & 0.30 & 0.45 \\
                Mild & 0.38 & 0.40 & 0.39 & 0.08 & 0.32 & 0.13 \\
                Moderate & 0.43 & 0.42 & 0.42 & 0 & 0 & 0 \\
                Severe & 0.44 & 0.44 & 0.44 & 0.38 & 0.48 & 0.43 \\
                \hline
                \hline
                \emph{Average} & 0.41 & 0.41 & 0.41 & 0.34 & 0.28 & 0.25 \\
                \cmidrule{2-4} \cmidrule{5-7}
                \emph{Overall Accuracy} & \multicolumn{3}{c}{0.41} & \multicolumn{3}{c}{0.34} \\
        \bottomrule
    \end{tabular}}
\end{table}

\begin{table}[h]
    \centering
    \caption{\textsc{Results for 2-Layer Depression Recognition Models.}} \label{tab:res-2}
    \resizebox{\textwidth}{!}{
    \begin{tabular}{ccccccc}
        \toprule
        \multirow{2}{*}{\textbf{Depression Level}} & \multicolumn{3}{c}{\textbf{2-layer+Xaiver}} & \multicolumn{3}{c}{\textbf{2-layer+Kaiming}} \\
                \cmidrule{2-4} \cmidrule{5-7}
                & \textbf{Precision} & \textbf{Recall} & \textbf{F1 score} & \textbf{Precision} & \textbf{Recall} & \textbf{F1 score} \\
        \midrule
                None & 0.55 & 0.53 & 0.54 & 0.51 & 0.48 & 0.50 \\
                Mild & 0.62 & 0.51 & 0.56 & 0.49 & 0.58 & 0.53 \\
                Moderate & 0.52 & 0.49 & 0.50 & 0.50 & 0.47 & 0.49 \\
                Severe & 0.45 & 0.67 & 0.54 & 0.57 & 0.56 & 0.57 \\
                \hline
                \hline
                \emph{Average} & 0.55 & 0.54 & 0.54 & 0.52 & 0.52 & 0.52 \\
                \cmidrule{2-4} \cmidrule{5-7}
                \emph{Overall Accuracy} & \multicolumn{3}{c}{0.54} & \multicolumn{3}{c}{0.52} \\
        \bottomrule
    \end{tabular}}
\end{table}

\begin{table}[h]
    \centering
    \caption{\textsc{Results for 3-Layer Depression Recognition Models.}} \label{tab:res-3}
    \resizebox{\textwidth}{!}{
    \begin{tabular}{ccccccc}
        \toprule
        \multirow{2}{*}{\textbf{Depression Level}} & \multicolumn{3}{c}{\textbf{3-layer+Xaiver}} & \multicolumn{3}{c}{\textbf{3-layer+Kaiming}} \\
                \cmidrule{2-4} \cmidrule{5-7}
                & \textbf{Precision} & \textbf{Recall} & \textbf{F1 score} & \textbf{Precision} & \textbf{Recall} & \textbf{F1 score} \\
        \midrule
                None & 0.54 & 0.52 & 0.53 & 0.74 & 0.78 & 0.76 \\
                Mild & 0.66 & 0.61 & 0.63 & 0.89 & 0.85 & 0.87 \\
                Moderate & 0.56 & 0.53 & 0.54 & 0.82 & 0.80 & 0.81 \\
                Severe & 0.56 & 0.68 & 0.61 & 0.87 & 0.89 & 0.88 \\
                \hline
                \hline
                \emph{Average} & 0.58 & 0.58 & 0.58 & 0.83 & 0.83 & 0.83 \\
                \cmidrule{2-4} \cmidrule{5-7}
                \emph{Overall Accuracy} & \multicolumn{3}{c}{0.58} & \multicolumn{3}{c}{0.83} \\
        \bottomrule
    \end{tabular}}
\end{table}

\subsection{Discussion}
According to the above results and analysis, although Kaiming initialization optimizes to the Xaiver method, it cannot outperform Xaiver in all situations. While an appropriate initialization method is important for boosting the performance of a model, which method to use depends on many important factors, such as the network topologies, hyperparameters, and the distribution of the dataset, etc. The results also enlighten us to focus more on the initialization methods to get better accuracy when using machine learning models in practice. 

Limited by the sample sizes, this paper does not involve further study on other important factors. For example, deeper neural networks are expected to perform better than the 3-layer configuration given more training samples. Besides, other network topologies such as CNNs and DNNs may also be applied. This inspires us to introduce and explore more appropriate network architectures in order to achieve higher prediction accuracy.

Nonetheless, the result reported in the paper demonstrates the effectiveness of gathering physical signals to training network models for depression prediction without human efforts, which is conducive to more objective diagnosis and early treatment of depression. Also, since the models learn the internal relationship between physical patterns and depression, we may also investigate what exactly the patterns that the model has learned, and which pattern and indicator is the most significant contributor to depression. They remain open questions to be answered in our future work.

%
%
%

\end{document}